\def\year{2022}\relax
\documentclass[letterpaper]{article} 
\usepackage{aaai22}  
\usepackage{times}  
\usepackage{helvet}  
\usepackage{courier}  
\usepackage[hyphens]{url}  
\usepackage{graphicx} 
\usepackage{natbib}  
\usepackage{caption} 
\DeclareCaptionStyle{ruled}{labelfont=normalfont,labelsep=colon,strut=off} 
\usepackage{algorithm}
\usepackage{algorithmic}
\usepackage{newfloat}
\usepackage{listings}
\usepackage{xcolor}
\usepackage{subcaption}

\floatstyle{ruled}
\newfloat{listing}{tb}{lst}{}
\floatname{listing}{Listing}
\title{MDD-Eval: Self-Training on Augmented Data for Multi-Domain Dialogue Evaluation}
\author {
    Chen Zhang\textsuperscript{\rm 1,2},
    Luis Fernando D’Haro\textsuperscript{\rm 4},
    Thomas Friedrichs\textsuperscript{\rm 2},
    Haizhou Li\textsuperscript{\rm 1,3,5}
}
\affiliations {
    \textsuperscript{\rm 1} National University of Singapore \quad 
    \textsuperscript{\rm 2} Robert Bosch (SEA), Singapore \quad  \textsuperscript{\rm 3} Kriston AI Lab, China\\
    \textsuperscript{\rm 4} Universidad Politécnica de Madrid, Spain \quad
    \textsuperscript{\rm 5} The Chinese University of Hong Kong (Shenzhen), China
     \\
    \{chen\_zhang,haizhou.li\}@u.nus.edu, luisfernando.dharo@upm.es, thomas.friedrichs@sg.bosch.com
}

\usepackage{booktabs}
\usepackage{amsmath}

\begin{document}

\maketitle

\begin{abstract}
Chatbots are designed to carry out human-like conversations across different domains, such as general chit-chat, knowledge exchange, and persona-grounded conversations. To measure the quality of such conversational agents, a dialogue evaluator is expected to conduct assessment across domains as well. However, most of the state-of-the-art automatic dialogue evaluation metrics (ADMs) are not designed for multi-domain evaluation. We are motivated to design a general and robust framework, MDD-Eval, to address the problem. Specifically, we first train a teacher evaluator with human-annotated data to acquire a rating skill to tell good dialogue responses from bad ones in a particular domain and then, adopt a self-training strategy to train a new evaluator with teacher-annotated multi-domain data, that helps the new evaluator to generalize across multiple domains. MDD-Eval is extensively assessed on six dialogue evaluation benchmarks. Empirical results show that the MDD-Eval framework achieves a strong performance with an absolute improvement of 7\% over the state-of-the-art ADMs in terms of mean Spearman correlation scores across all the evaluation benchmarks.   
\end{abstract} 




\section{Introduction}
\label{sec:intro}

Recent years have witnessed growing interests in open-domain dialogue systems~\citep{adiwardana2020towards,zhang-etal-2020-dialogpt,roller-etal-2021-recipes}. With the increasing availability of high-quality dialogue corpora~\citep{li-etal-2017-dailydialog,zhang-etal-2018-personalizing} and advancement of neural architectures~\citep{devlin2019bert,radford2019language}, learning-based dialogue systems are becoming possible. The applications call for  dialogue technology capable of generating appropriate responses to users' prompts in a diverse range of scenarios, such as general chit-chat~\citep{li-etal-2017-dailydialog}, knowledge exchange~\citep{gopalakrishnan2019topical}, persona-based chat~\citep{zhang-etal-2018-personalizing}, and emotion disclosure~\citep{rashkin-etal-2019-towards}.  

However, the dialogue research heavily relies on the ability to evaluate system performance with automatic dialogue evaluation metrics (ADMs). Common natural language generation (NLG) metrics used in the dialogue system literature, such as BLEU~\citep{papineni2002bleu} and ROUGE~\citep{lin-2004-rouge}, are unsuitable for the multi-domain dialogue evaluation task as they are shown to correlate poorly with human judgements~\citep{liu-etal-2016-evaluate} due to the one-to-many context-response mapping in dialogues~\citep{zhao-etal-2017-learning} as well as the multi-faceted nature of dialogue evaluation~\citep{mehri-eskenazi-2020-usr}. 



\begin{table}[!t]
\centering
\resizebox{0.8\linewidth}{!}{
\begin{tabular}{@{}cccccc@{}}
\toprule
\multicolumn{2}{l}{Metric} & \multicolumn{2}{c}{DailyDialog-Eval} & \multicolumn{2}{c}{Topical-Eval} \\ \midrule
\multicolumn{2}{l}{DEB} & \multicolumn{2}{c}{0.486} &  \multicolumn{2}{c}{0.116} \\
\multicolumn{2}{l}{GRADE} & \multicolumn{2}{c}{0.533} & \multicolumn{2}{c}{0.217} \\
\multicolumn{2}{l}{USR} & \multicolumn{2}{c}{0.367} &\multicolumn{2}{c}{0.423} \\ \bottomrule
\end{tabular}
}
\caption{Spearman correlation scores of three state-of-the-art model-based metrics on two dialogue evaluation benchmarks.}
\label{tab:tab1}
\end{table}

An alternative solution is to design model-based ADMs that explicitly learn to discriminate dialogue responses of varying quality. Lately, many model-based ADMs leveraging self-supervised learning are proposed to address the weaknesses of the standard NLG metrics~\citep{sai-etal-2020-improving,ghazarian-etal-2019-better,mehri-eskenazi-2020-usr,huang-etal-2020-grade,zhang-etal-2021-dscore}. While these ADMs have demonstrated strong correlations with human judgements, they lack a generalized skill to evaluate dialogues across multiple domains. For example, in Table~\ref{tab:tab1}, DEB~\citep{sai-etal-2020-improving} and GRADE~\citep{huang-etal-2020-grade} are pretrained on the DailyDialog dataset~\citep{li-etal-2017-dailydialog}. They perform well on the DailyDialog-Eval~\citep{zhao-etal-2020-designing} benchmark that contains responses from dialogue systems trained on chit-chat content. However, their performance significantly drops when assessed on the Topical-Eval~\citep{mehri-eskenazi-2020-usr} benchmark, which is close in domain with TopicalChat~\citep{gopalakrishnan2019topical} and contains dialogue responses from knowledge-grounded conversations. The reverse is true for USR~\citep{mehri-eskenazi-2020-usr}, which is pretrained on the TopicalChat dataset.

To design robust ADMs for the multi-domain dialogue evaluation task, we consider two research questions. (1) How to equip the ADM with a rating skill to discriminate responses of varying quality? In other words, the ability to assign a high score to relevant responses and a low score otherwise. (2) How can an ADM learn the general knowledge across dialogue domains so as to generalize the evaluation skill? For the first question, the most direct and effective way is to learn from humans, i.e., the ADM can be trained with human-annotated dialogue data. As for the second question, the general knowledge can be learned on a large-scale multi-domain dialogue dataset. Ideally, If human annotations are available, an oracle multi-domain dialogue evaluator can be learned. However, performing large-scale human annotations is extremely expensive. Thus, we are motivated to explore semi-supervised learning for our task. 




More specifically, we propose a multi-domain dialogue evaluation (MDD-Eval) framework under the self-training paradigm~\citep{scudder1965probability,yarowsky1995unsupervised} where a teacher model, trained on human-annotated dialogue evaluation data, creates pseudo labels for unlabeled dialogue data. Then, the synthetically-labeled data are used to train a student model. To obtain the large-scale multi-domain unlabeled dialogue data, we leverage the dialogue data augmentation techniques that have been successfully applied in the self-supervised learning of ADMs, such as random utterance selection~\citep{tao2018ruber,zhang-etal-2021-dscore}, mask-and-fill~\citep{donahue-etal-2020-enabling,gupta-etal-2021-synthesizing} and back-translation~\citep{edunov-etal-2018-understanding,sinha-etal-2020-learning}. In this way, we expect that the student model carries the rating skill of the teacher model, and it can generalize across domains after being adapted on a large-scale multi-domain dataset with pseudo labels.


\bigskip
\noindent Overall, we make the following contributions:
                                     
\begin{itemize}
    \item A model-based framework, named MDD-Eval, is proposed with a self-training scheme on augmented data. Its rating skill is trained on human-annotated data, and its cross-domain general knowledge is trained on machine-annotated data. 
    \item We release a large-scale multi-domain dialogue dataset with machine annotations that facilitate ADM training. We name the dataset, MDD-Data.  
    \item 
    MDD-Eval attains an absolute improvement of 7\% over the state-of-the-art ADMs in terms of mean Spearman correlation over six dialogue evaluation benchmarks.
    
    
    
    \item MDD-Data, MDD-Eval implementation, and pretrained checkpoints will be released to the public\footnote{https://github.com/e0397123/MDD-Eval}. This allows practitioners and researchers to use and adapt MDD-Eval for automatic evaluation of their dialogue systems.  
\end{itemize}

\section{Related Work}
\subsection{Dialogue Evaluation Metrics}
Human evaluation reflects the perceived quality of dialogue systems. However, it is expensive and time-consuming. For system development, we rely on ADMs for model design, hyperparameter tuning and system benchmarking~\citep{yeh2021comprehensive}. The current trend of open-domain ADMs is shifting from the reference-based approach towards the model-based approach that is reference-free~\citep{mehri-eskenazi-2020-unsupervised,zhang-etal-2021-dynaeval}. In many ADM solutions, we predict the relatedness between a dialogue context and the generated responses by training a discriminative network to distinguish the original response from negative samples in a self-supervised fashion. Typical examples include RUBER~\citep{tao2018ruber}, BERT-RUBER\citep{ghazarian-etal-2019-better}, USR~\citep{mehri-eskenazi-2020-usr}, GRADE~\citep{huang-etal-2020-grade}, MaUdE~\citep{sinha-etal-2020-learning} and D-score~\citep{zhang-etal-2021-dscore}. 

A problem with the metrics learned with self-supervised learning is that the random negative-sampling strategy is likely to produce false-negative or over-simplistic candidates, thus introducing unwanted biases to the ADMs. One idea is to introduce adversarial irrelevant responses to increase the ADMs' discrimination capability~\citep{sai-etal-2020-improving,gupta-etal-2021-synthesizing,park-etal-2021-generating}. In this way, the evaluation model will greatly benefit from a dataset of multiple relevant and adversarial irrelevant responses from diverse dialogue context. The existing methods are focused very much on how to design such a dataset. Along this line of thought, this work presents a novel strategy to learn the rating skill from one dataset first, then generalize the skill across multiple domains. 

\subsection{Self-Training}
Self-training is a simple and effective semi-supervised approach, which incorporates a model’s prediction on unlabeled data to obtain additional information. It has been shown effective in many tasks, such as image recognition~\citep{yalniz2019billion}, text generation~\citep{he2019revisiting}, automatic speech recognition~\citep{kahn2020self}, and parsing~\citep{mcclosky-etal-2006-effective}. There are two key ideas that contribute to the success of self-training: pseudo-labeling and consistency regularization. 

Pseudo-labeling refers to the process of converting model predictions to hard labels~\citep{lee2013pseudo}. Usually, a confidence-based threshold is imposed to retain unlabeled examples only when the classifier is sufficiently confident~\citep{sohn2020fixmatch}. In MDD-Eval, we apply pseudo-labeling together with the confidence-based threshold to
bootstrap high-quality adversarial and random negative samples from the unlabeled data.

Consistency regularization was first proposed by~\citep{bachman2014learning}. It means that the prediction made by the classification model remains consistent even when the input or the model function is perturbed by a small amount of noise. Recently, the use of consistency regularization to modulate the self-training process has been shown to boost model performance on many image and text classification tasks~\citep{xie2020unsupervised,Berthelot2020ReMixMatch}. We are motivated to incorporate consistency regularization into the learning of our dialogue evaluator, which is essentially learned with a text classification task.

~\citet{xie2020self} proposes Noisy Student and~\citet{sohn2020fixmatch} proposes FixMatch frameworks. Both incorporate pseudo-labeling and consistency regularization into a unified framework.  Noisy Student and FixMatch have demonstrated remarkable performance on image classification tasks, that motivates us to unify the pseudo-labeling and consistency regularization ideas in open-domain ADM training for the first time.







\section{Methodology}
\label{sec:method}
In this section, we first define the multi-domain dialogue evaluation task (Section~\ref{subsec:problem-formulation}), then formulate MDD-Eval framework in three steps: (a) We pretrain a teacher model (Section~\ref{subsec:teacher-model}) from a human-annotated dataset, to learn the rating skill to distinguish relevant responses from irrelevant ones. (b) We augment a large-scale multi-domain dataset for MDD-Eval self-training (Section~\ref{subsec:dialog-augmentation}). (c) We generalize the pretrained teacher model with the augmented data to derive a student model, which carries a generalized rating skill learned from the augmented data. (Section~\ref{subsec:student-model}). 

\subsection{Problem Formulation}
\label{subsec:problem-formulation}
Formally, a dialogue context and the corresponding dialogue response can be denoted as $c_i^j$ and $r_i^j$ respectively. $c_i^j$ and $r_i^j$ are the $i^{th}$ data pair drawn from the $j^{th}$ dialogue evaluation benchmark $D^j$, where $j\in\{1,...,J\}$, and $D^j \in D^J$ and $i\in\{1,...,I\}$. There are $J$ domains, each of which has $I$ data pairs. 


Our goal is to learn a metric, $M: (c_i^j, r_i^j) \rightarrow s_i^j$ where $s_i^j$ is the metric score that indicates the quality of $(c_i^j, r_i^j)$ as perceived by $M$. In addition, each $(c_i^j, r_i^j)$ is annotated by several human judges and each human judge will provide a quality score based on the Likert scale\footnote{In the evaluation benchmarks used in our experiments, the Likert scale is from 1 to 5. The higher the better.} to indicate his or her perception of the quality of $(c_i^j, r_i^j)$. We denote the mean human score given to $(c_i^j, r_i^j)$ as $q_i^j$. Due to the multi-faceted nature of dialogue evaluation, the quality can refer to language fluency, coherence, topic relevance, logical consistency etc. Since the focus of our work is multi-domain dialogue evaluation instead of multi-dimensional evaluation, we fix the quality as \textit{response appropriateness} here. 

To assess the performance of $M$ on $D^j$, the correlation score between $S = \{s_i^j,\ldots,s_I^j\}$ and $Q = \{q_i^j,\ldots,q_I^j\}$ are calculated. We use $\rho_j$ to represent the correlation score on $D^j$. Higher $\rho_j$ indicates better performance of the metric on $D^j$. In the multi-domain dialogue evaluation task, an effective $M$ should achieve good correlation scores across all ${J}$ domains. In other words, the desired $M$ should obtain a good average correlation $\tilde{\rho} = \frac{1}{J}\sum^{J}_{j=1}{\rho_j}$.

\subsection{Teacher Model}
\label{subsec:teacher-model}
We first pretrain a model on human-annotated data in one particular domain, i.e., the teacher model, $M_{teacher}$, defined by the parameters   $\theta_{teacher}$. Given a dialogue context-response pair, $M_{teacher}$ should accurately determine the degree of relevance between the context and the corresponding response. To equip the teacher model with a solid rating skill, we rely on a high-quality human-annotated base dataset $D^b\in{D^J}$. Note that $D^b$ is from a single-domain, and of much smaller size than the data we would like to augment. 

In dataset $D^b$, there are three categories of responses for a given context: random, adversarial and relevant. The relevant and adversarial responses are generated by human annotators. $M_{teacher}$ is trained on $D^b$ to classify a context-response pair into one of the three categories:
\begin{equation}
\tilde{y}_i^b = f_{\theta_{teacher}}([c_i^b \circ r_i^b])
\label{eq:teacher_prediction}
\end{equation}
with the objective function:
\begin{equation}
    \underset{\theta_{teacher}}{\min}\frac{1}{|D^b|}\sum_{(c_i^b, r_i^b, y_i^b)\in{D^b}}{\mathcal{L}_{CE}(\tilde{y}_i^b, y_i^b)}
\end{equation}
where $\circ$ denotes the concatenation operation. $\tilde{y}_i^b$ is the predicted class, $y_i^b$ is the gold label for $(c_i^b, r_i^b)$ and $\mathcal{L}_{CE}$ is the cross entropy loss. 

\begin{figure}[t!]
\centering
\includegraphics[width=0.4\textwidth]{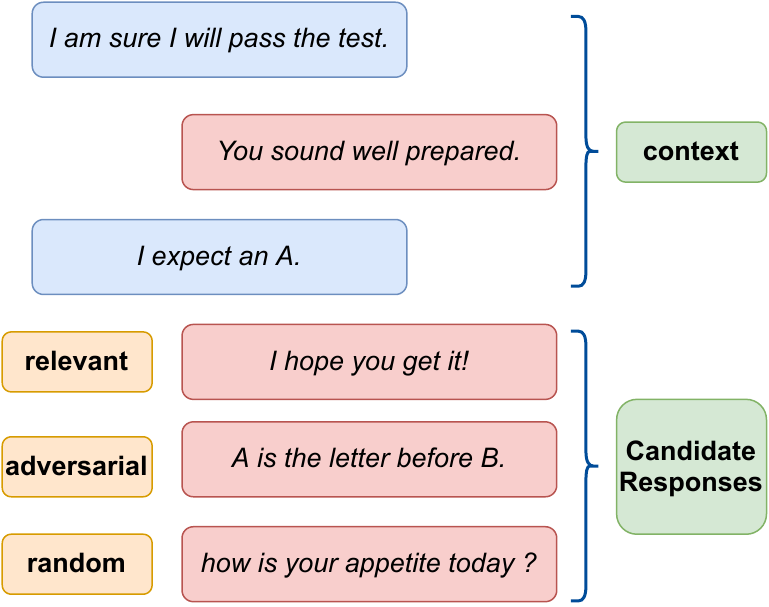} 
\caption{An example of a dialogue context with three candidate responses. $M_{teacher}$ is expected to annotate the context-response pairs as either relevant, adversarial or random.}  
\label{context-response-example}
\end{figure}

$M_{teacher}$ plays three key roles: (1) providing pseudo labels to unlabeled context-response pairs, $(c_i^{\text{*}}, r_i^{\text{*}})$\footnote{$*$ means that the context-response pair can be drawn from dialogue corpora of any domain.}, which are obtained with different dialogue data augmentation techniques. (2) facilitating the data selection process whereby false negatives and adversarial or random samples with low confidence scores as determined by $M_{teacher}$ are removed. (3) serving as a baseline in the evaluation task.

\subsection{Dialogue Data Augmentation}
\label{subsec:dialog-augmentation}

To generalize the teacher model across domains, we collect a multi-domain dataset, denoted as $D^{\text{*}}$, that contains a large amount of unlabeled context-response pairs. The unlabeled pairs will be automatically annotated in the same way as $D^b$ by $M_{teacher}$. An example of a dialogue context with three candidate responses for annotation is presented in Figure~\ref{context-response-example}. To construct such a dataset, we leverage the following dialogue data augmentation techniques:
\\

\noindent\textbf{Syntactic Perturbation}
Motivated by~\citep{sinha-etal-2020-learning}, we have considered three variants of perturbations at the syntax level: (1) word-drop (a random portion of tokens in the response is dropped). (2) word-shuffle (the ordering of tokens in the response is randomly shuffled). (3) word-repeat (a random portion of tokens in the response is repeated multiple times). The syntactic perturbations are intended to simulate erroneous behaviours of some generative models in generating unnatural dialogue responses.
\\

\noindent\textbf{Back-Translation}
Back-translation~\citep{edunov-etal-2018-understanding} augments a response by generating its syntactic variants. In practice, we adopt the pretrained WMT’19 English-German and German-English ensemble model to perform back-translation. 
\\  

\noindent\textbf{Generative Model Output}
State-of-the-art dialogue generators, such as DialoGPT~\citep{zhang-etal-2020-dialogpt} and BlenderBot~\citep{roller-etal-2021-recipes}, have been pretrained on a large amount of conversation data and are demonstrating strong capability in generating fluent and on-topic responses. They help generate semantic variants of a response conditioned on the respective dialogue contexts.  
\\

\noindent\textbf{Random Utterance Selection}
The random utterance selection is a simple and effective strategy that has been widely adopted in the self-supervised learning of dialogue evaluation metrics~\citep{mehri-eskenazi-2020-usr,huang-etal-2020-grade,sai-etal-2020-improving} to introduce irrelevant responses w.r.t. a dialogue context. Given a dialogue context, three variants of random utterance selection are adopted: (1) randomly sample a response from a different dialogue. (2) randomly sample a response from the entire pool of responses produced by the generative models. (3) randomly sample a response from the entire pool of responses obtained via back-translation.
\\

\noindent\textbf{Mask-and-fill}
Above-mentioned techniques tend to produce response candidates for the relevant and random class. The mask-and-fill strategy is adopted to automatically construct candidates for the adversarial class. Specifically, we adopt the Infilling by Language Modeling (ILM) framework~\citep{donahue-etal-2020-enabling} to perform the mask-and-fill response augmentation. The process is as follows: given a context-response pair extracted from a natural human-human dialogue, one or a few contiguous tokens in the response are randomly replaced by the $[blank]$ placeholder. The modified response is input into the pretrained ILM model, which then generate tokens in an autoregressive manner. Subsequently, the $[blank]$ placeholder is substituted with the generated tokens to obtain a reconstructed view of the original response. The reconstructed response serves as an adversarial sample w.r.t. the dialogue context.
\\

\noindent
After obtaining the large number of context-response pairs, we apply the pretrained $M_{teacher}$ to provide soft pseudo labels to all the pairs. The soft pseudo label is a probability distribution over the three classes (random, adversarial and relevant). Then, a filtering process is implemented to improve the quality of pseudo-labeled $D^{\text{*}}$. A confidence threshold of 70\% is applied to exclude pairs classified by $M_{teacher}$ with low confidence. Emprical evidence suggests that the 70\% threshold provides a good balance between the quality and quantity of augmented data. Within $D^{\text{*}}$, the relevant set consists of filtered pairs obtained with back-translation and generative models in addition to the original context-response pairs extracted from dialogues of different dialogue corpora. The adversarial set mainly include filtered pairs that are constructed via syntactic perturbation and mask-and-fill strategy. For the random set, the context-response pairs are mainly obtained with random utterance selection. 

\begin{figure}[t]
\centering
\includegraphics[scale=0.45]{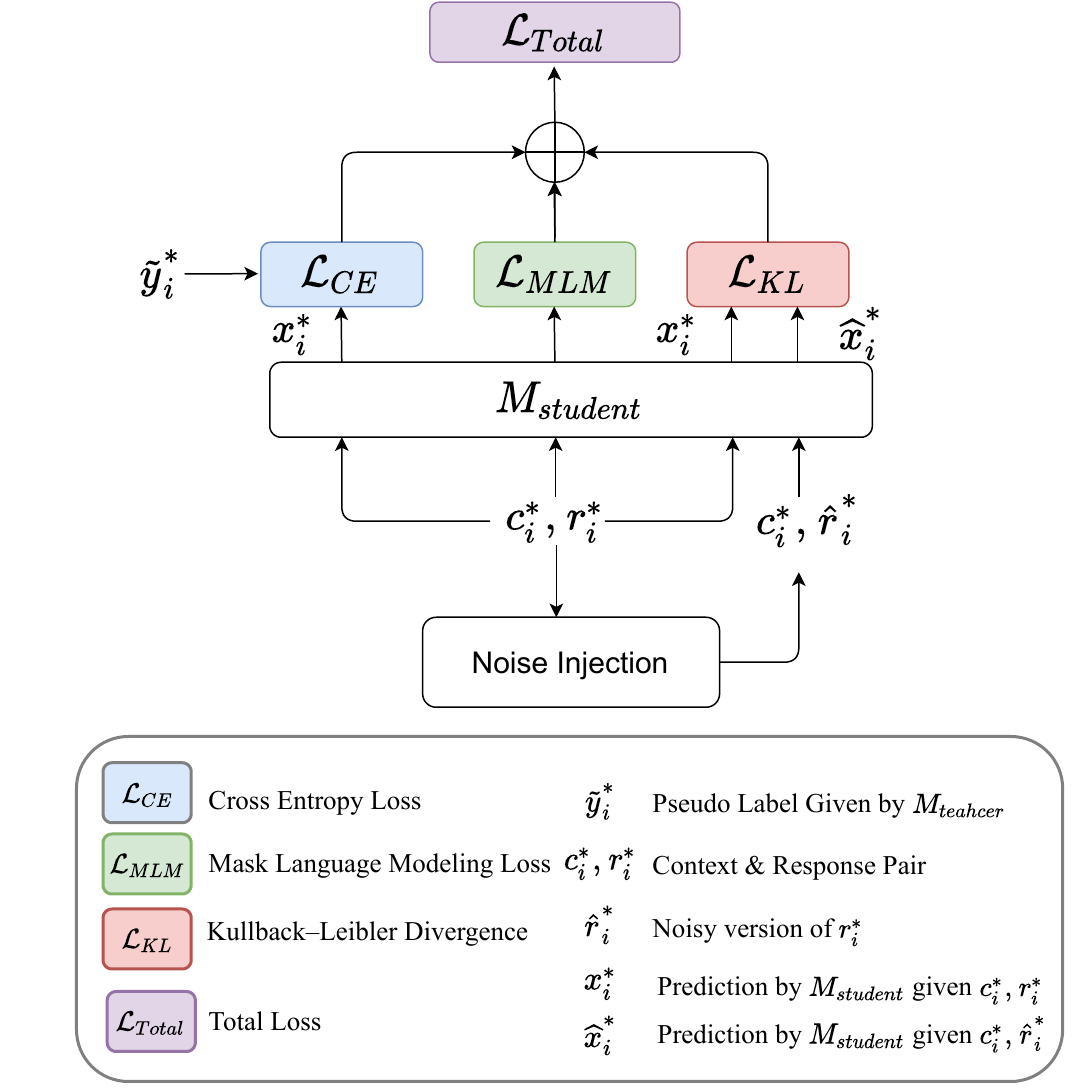} 
\caption{The training process of $M_{student}$. $\mathcal{L}_{Total}$ is the sum of three components: (1) The cross entropy loss $\mathcal{L}_{CE}$, which is computed between $\tilde{y}_i^{*}$ generated by $M_{teacher}$ and the prediction by $M_{student}$ for an input pair $(c_i^{*},r_i^{*})$ . (2) The self-supervised MLM loss $\mathcal{L}_{MLM}$, for domain adaptation. (3) The KL Loss $\mathcal{L}_{KL}$, for consistency regularization.}    

\label{student-model-objective}
\end{figure}

\subsection{Student Model}
\label{subsec:student-model}
Once $D^{\text{*}}$ is ready, we can learn a student model, $M_{student}$ parameterized by $\theta_{student}$, on $D^{\text{*}}$ by performing the following classification task:
\begin{equation}
x_i^{\text{*}} = f_{\theta_{student}}([c_i^{\text{*}} \circ r_i^{\text{*}}])
\label{eq:student_prediction}
\end{equation}
Figure~\ref{student-model-objective} is a graphical illustration of the training objective of $M_{student}$ and the equation is as follows: 
\begin{equation}
\begin{aligned}
& \underset{\theta_{student}}{\min}\frac{1}{|D^{\text{*}}|} \sum_{(c_i^{\text{*}}, r_i^{\text{*}}, \tilde{y}_i^{\text{*}})\in{D^{\text{*}}}}{\mathcal{L}_{CE}(x_i^{\text{*}}, \tilde{y}_i^{\text{*}})} +  \\
& \mathcal{L}_{KL}(x_i^{\text{*}}, \hat{x}_i^{\text{*}}) + \mathcal{L}_{MLM}([c_i^{\text{*}} \circ r_i^{\text{*}}]) \\
\end{aligned}
\label{eq:4}
\end{equation}
where $\mathcal{L}_{CE}$ is the cross-entropy loss, $\mathcal{L}_{KL}$ is the KL divergence and $\mathcal{L}_{MLM}$ is the self-supervised masked language modeling (MLM) loss. $x_i^{\text{*}}$ and $\tilde{y}_i^{\text{*}}$ are the logits output from $M_{student}$ and the pseudo label generated by the pretrained $M_{teacher}$ respectively given the input pair $(c_i^{\text{*}}, r_i^{\text{*}})$.

$\mathcal{L}_{KL}$ is introduced to enforce consistency regularization, with which $M_{student}$ is less sensitive to noise and hence smoother w.r.t. perturbations in the input space~\citep{xie2020unsupervised}. We denote the noisy version of $r_i^{\text{*}}$ after noise injection as $\hat{r}_i^{\text{*}}$. In the practical implementation, we follow~\citep{he2019revisiting} to generate $\hat{r}_i^{\text{*}}$ based on $r_i^{\text{*}}$. $\hat{x}_i^{\text{*}}$ is the corresponding logits from $M_{student}$ after inputting $(c_i^{\text{*}}, \hat{r}_i^{\text{*}})$. The KL divergence between the respective post-softmax probability distributions of $x_i^{\text{*}}$ and $\hat{x}_i^{\text{*}}$ is minimized during training.

The last term, $\mathcal{L}_{MLM}$, is intended to help $M_{student}$ extract additional domain-specific knowledge so as to better adapt to the multi-domain synthetic dataset. The MLM implementation follows the standard BERT~\citep{devlin2019bert} practice whereby a random portion of tokens in the concatenated sequence, $[c_i^{\text{*}} \circ r_i^{\text{*}}]$, are masked. $M_{student}$ is expected to make predictions on the masked tokens.


\subsection{Run-time Scoring Process}
\label{subsec:score-process}
The learned student model serves as the backbone of MDD-Eval for performing the multi-domain dialogue evaluation task, that derives the metric score $s_i^j$ for a given context-response pair $(c_i^j, r_i^j)\in{D^j}$ as mentioned in Section~\ref{subsec:problem-formulation}. We formulate the scoring process by $M_{student}$ as follows:
\begin{equation}
s_i^j = P(\tilde{y}_i^j=\textrm{relevance}|(c_i^j,r_i^j))
\end{equation}
which is the post-softmax probability w.r.t. the relevant class output by $M_{student}$ given the input, $(c_i^j,r_i^j)$.  


\section{Experiment Setup}
\label{sec:experiment}
We first discuss the dialogue corpora (Section~\ref{subsec:datasets}) used in the experiments. Then, the evaluation benchmarks used for assessing the performance of MDD-Eval are discussed in Section~\ref{subsec:benchmarks}. Next, Section~\ref{subsec:architecture} is about the architecture choice for both the teacher and the student model. Finally, the choices of baselines are outlined in Section~\ref{subsec:baselines}.    

\subsection{Dialogue Corpora}
\label{subsec:datasets}
\bigskip
\noindent\textbf{Base for Teacher Training}
DailyDialog++~\citep{sai-etal-2020-improving} is a multi-reference dialogue evaluation dataset developed based on DailyDialog~\citep{li-etal-2017-dailydialog}; In this work, it is selected as the base dataset. In total, DailyDialog++ contains 11,429 dialogue contexts and the average number of turns per context is 3.31. There are three categories of responses: random, adversarial and relevant. For each context, the authors collected five different responses per category. Both the relevant and adversarial responses are written by human annotators. The adversarial responses share certain degree of lexical or semantic overlap with the corresponding dialogue contexts, but are still deemed as inappropriate responses. They are introduced to avoid model decision-making based on spurious features in the context-response pairs.

\begin{table*}[!h]
\centering
\resizebox{0.8\textwidth}{!}{
\begin{tabular}{l|cc|l|cc}
\toprule
\textbf{DailyDialog} & \textbf{training} & \textbf{validation} & \textbf{EmpatheticDialog} & \textbf{training} & \textbf{validation} \\ \midrule
\#dialogues & 11,118 & 1,000 & \#dialogues & 19,529 & 2,768  \\
\#utterances & 87,170 & 8,069 & \#utterances & 84,158 & 12,075 \\
\#words & 1,186,046  & 108,933 & \#words & 1,127,355 & 174,786  \\
\#avg utterances per dialogue & 7.84 & 8.07 & \#avg utterances per dialogue & 4.31  & 4.36 \\ 
\#avg words per dialogue & 106.68 & 108.93 & \#avg words per dialogue & 57.73  & 63.15 \\  \midrule
\textbf{ConvAI2} & \textbf{training} & \textbf{validation} & \textbf{TopicalChat} & \textbf{training} & \textbf{validation} \\ \midrule
\#dialogues & 17,878 & 1000 & \#dialogues & 8,627 & 538 \\
\#utterances & 253,698 & 15,566 & \#utterances & 188,357  & 11,660 \\
\#words & 3,024,032  & 189,374 & \#words & 4,374,304 &  273,331 \\
\#avg utterances per dialogue & 14.19 & 15.57 & \#avg utterances per dialogue & 21.83 & 21.67 \\ 
\#avg words per dialogue & 169.15 & 189.37 & \#avg words per dialogue & 507.05 & 508.05 \\
\bottomrule
\end{tabular}
}
\caption{Human-Human Dialogue Corpora Statistics}\label{tab:data-statistics}
\end{table*}

\bigskip
\noindent\textbf{Multi-domain Dialogue Corpora for Augmentation}
We make use of four publicly-available, high-quality and human-written conversation corpora to form a multi-domain synthetic dataset: DailyDialog~\citep{li-etal-2017-dailydialog}, ConvAI2~\citep{dinan2020second}, EmpatheticDialogues~\citep{rashkin-etal-2019-towards} and TopicalChat~\citep{gopalakrishnan2019topical}. The detailed statistics of the four dialogue corpora are presented in Table~\ref{tab:data-statistics}.  We only use the training and validation sets of the dialogue corpora since some dialogue contexts in the evaluation benchmarks are sampled from their test sets.

To extract context-response pairs from the human-human dialogues, we take the dialogue history and current response as an original context-response pair. The number of utterances per context is kept between one and four. For each original context-response pair, we sample ten different augmented pairs per augmentation technique. After the filtering process, we end up with a class-balanced and multi-domain synthetic dataset of around 2.6 million context-response pairs. We name this synthetic dataset, MDD-Data. In our experiment, for quick turn-around, we sub-sample 600K context-response pairs from MDD-Data to train the final student model. 

\subsection{Evaluation Datasets}
\label{subsec:benchmarks}
Guided by~\citep{yeh2021comprehensive}, we use publicly-available dialogue evaluation datasets to assess the performance of the ADMs. Additionally, we propose a few criteria for the selection of high-quality dialogue evaluation datasets. First, we select the ones that cover as many domains as possible. Second, the size of the datasets should be sufficiently large to provide statistically significant analysis. Third, most of the state-of-the-art metrics should have achieved relatively good correlation results on the datasets. This is to avoid inclusion of any biased evaluation dataset. Next, the inter-annotator agreement should be relatively good ($\sim$0.6). Lastly, the evaluation datasets should cover responses of a wide quality spectrum. In total, we have adopted six different publicly-available dialogue evaluation datasets with each accounting for a dialogue domain for assessing MDD-Eval\footnote{The names of the datasets are unified in our paper to better distinguish their respective domains.}: DailyDialog-Eval~\citep{zhao-etal-2020-designing},  Persona-Eval~\citep{zhao-etal-2020-designing}, Topical-Eval~\citep{mehri-eskenazi-2020-usr}, Movie-Eval~\citep{app10030762}, Empathetic-Eval~\citep{huang-etal-2020-grade} and Twitter-Eval~\citep{hori2017end}. Detailed statistics of each evaluation dataset is listed in Table~\ref{tab:benchmar-statistics}.

\begin{table*}[!h]
\centering
\resizebox{\textwidth}{!}{
    \begin{tabular}{l|ccccccl}
    \toprule
    Name & \#Instances & Avg.\#Utts. & Avg.\#Ctx/Hyp Words & Type & \#Criteria & \#Annotations & Used NLG models\\
    \midrule
    Persona-Eval~\shortcite{zhao-etal-2020-designing} & 900 & 5.1 & 48.8 / 11.5 & Turn-level & 1 & 3,600 & LSTM Seq2Seq, Random sampling, and GPT-2\\
    DailyDialog-Eval~\shortcite{zhao-etal-2020-designing} & 900 & 4.7 & 47.5 / 11.0 & Turn-level & 4 & 14,400 & LSTM Seq2Seq, Random sampling, and GPT-2\\
    Topical-Eval~\shortcite{mehri-eskenazi-2020-usr} & 360 & 11.2 & 236.3 / 22.4 &  Turn-level & 6 & 6,480 & Transformers\\
    Empathetic-Eval~\shortcite{huang-etal-2020-grade} & 300 & 3.0 & 29.0 / 15.6 & Turn-level & 1 & 3,000 & Transformer Seq2Seq, Transformer Ranker \\
    Twitter-Eval~\shortcite{hori2017end} & 9,990 & 3.5 & 35.3 / 11.2 & Turn-level & 3 & 29,700 & RNN, LSTM Seq2Seq \\
    Movie-Eval~\shortcite{app10030762} & 9,500 & 3.9 & 17.0 / 6.1  & Turn-level & 2 & 57,000 & Random sampling\\
    \bottomrule
    \end{tabular}
}
    \caption{Summary of the evaluation datasets. Some information are obtained from~\citep{yeh2021comprehensive} and~\citep{chen2021automatic}. \#criteria is the number of response qualities that have been annotated, such as appropriateness, naturalness, etc.}
    \label{tab:benchmar-statistics}
\end{table*}

\subsection{Model Architecture Choice}
\label{subsec:architecture}
We choose RoBERTa-Large~\citep{liu2019roberta} for both the teacher and the student model in MDD-Eval. There are two reasons. First, RoBERTa has been pretrained on more than 160GB of uncompressed text covering multiple domains including news, stories, books and web text. Therefore, it equips the prediction model with general knowledge of the text with which the prediction model can easily adapt to the downstream dialogue evaluation tasks. Second, it has been proven as a powerful text encoder that are beneficial for the automatic dialogue evaluation task in prior works~\citep{zhao-etal-2020-designing,mehri-eskenazi-2020-usr,zhang-etal-2021-dscore,zhang2021investigating}.

\subsection{Baselines}
\label{subsec:baselines}
We compare MDD-Eval against state-of-the-art reference-free dialogue metrics, including DEB~\citep{sai-etal-2020-improving}, USL-H~\citep{phy-etal-2020-deconstruct}, GRADE~\citep{huang-etal-2020-grade},  USR~\citep{mehri-eskenazi-2020-usr}, unreferenced BERT-RUBER (uBERT-R)~\citep{ghazarian-etal-2019-better}, and D-score~\citep{zhang-etal-2021-dscore}. The selection of baselines is guided by a recent comprehensive survey on ADMs~\citep{yeh2021comprehensive}, which has showcased the strong performance of the above-mentioned metrics. In fact, each selected metric is one of the top-ranking metrics on one or more public dialogue evaluation benchmarks.  As in the previous work, we use the publicly-available checkpoints of the selected evaluation metrics for our evaluation tasks. Table~\ref{tab:metric-training-details} summarizes the training details of the model-based evaluation metrics including the teacher (MDD-T) and student models (MDD-S) in MDD-Eval as well.

\begin{table*}[!t]
\centering
\resizebox{\linewidth}{!}{
\begin{tabular}{@{}lcccccc@{}}
\toprule
& Training Dataset & Size & Pretrained Model & Objective & External Knowledge & Single
\\ \midrule
DEB & DailyDialog++ & $\sim$139K & BERT & CrossEntropy & Reddit Conversations & Yes\\
GRADE & DailyDialog & $\sim$178K & BERT & Triplet & ConceptNet & Yes \\
USR & TopicalChat / PersonaChat  & Unknown & RoBERTa & MLM / CrossEntropy & Persona Profiles / Wikipedia  & No \\
D-score & PersonaChat / Twitter / DailyDialog & $\sim$1.31M / $\sim$1.16M & RoBERTa & MLM / CrossEntropy & None & No \\ 
USL-H & DailyDialog & $\sim$138K & BERT & MLM / CrossEntropy & None & No \\
uBERT-R & DailyDialog / PersonaChat & Unknown & BERT & Triplet & None & Yes \\
MDD-T & DailyDialog++ & $\sim$139K & RoBERTa & CrossEntropy & None & Yes \\
MDD-S & MDD-Data & $\sim$600K & RoBERTa & CrossEntropy/ MLM / KL  & None & Yes \\
\bottomrule
\end{tabular}
}
\caption{Training details of model-based metrics. ‘Training Dataset’ and ‘Size’ indicate the training dialogue corpora and training data size in terms of the number of context-response pairs respectively. ‘Pretrained Model’ and ‘Objective’ refer to the backbone pretrained language model and the loss function used by the metrics accordingly. ‘External Knowledge’ means whether the training process leverages additional knowledge sources. ‘Single’ denotes that whether a metric is a single evaluation model or a combination of multiple evaluation models. ‘Unknown’ means that information is not publicly available.}
\label{tab:metric-training-details}
\end{table*}

\section{Results \& Analysis}
\label{sec:analysis}


\begin{table*}[!ht]	
\centering
\small
\resizebox{0.8\linewidth}{!}{
\begin{tabular}{@{}lcccccc|ccc|c@{}}
\toprule
\multicolumn{7}{c}{\textbf{Baselines}} & \multicolumn{3}{c}{\textbf{Ablation Metrics}} & \multicolumn{1}{c}{\textbf{Final}} \\
\midrule
\textbf{Benchmarks}  & \textbf{DEB} & \textbf{USL-H} & \textbf{GRADE} & \textbf{USR} & \textbf{uBERT-R} &  \textbf{D-score} & \textbf{MDD-T} & \textbf{MDD-C} & \textbf{MDD-CM} & \textbf{MDD-S} \\ \midrule
DailyDialog-Eval & 0.486 & 0.391  & 0.533 & 0.367  & 0.285 & 0.426 & 0.501 & 0.482 & 0.546 &\textbf{0.579}\\
Persona-Eval & 0.579 & 0.407 & 0.583 & 0.571 & 0.384 & 0.511 & 0.528 & 0.580 & 0.594 &\textbf{0.621}\\
Topical-Eval & 0.116  & 0.340 &  0.217 & 0.423 & 0.348 & 0.233 & 0.218 & 0.373 & 0.484 & \textbf{0.520}\\
Empathetic-Eval & 0.395 & 0.235  & 0.297 & 0.255 & 0.148 & \underline{0.087} & 0.345 & 0.404 & \textbf{0.404} & 0.374 \\
Movie-Eval & \textbf{0.649} & 0.531 & 0.612 & 0.366 & 0.388 & 0.340 & 0.383 & 0.556  & 0.524 & 0.537 \\
Twitter-Eval & 0.214  & 0.179 & 0.122 & 0.166 & 0.217 & \textbf{0.301} & 0.249 & 0.258 & 0.241 & 0.227 \\ \midrule
Average & 0.407 & 0.347 & 0.394 & 0.358 & 0.295 & 0.316 & 0.371 & 0.442 & 0.466 & \textbf{0.476}\\ \bottomrule
\end{tabular}}
\caption{Spearman correlation scores of state-of-the-art ADMs and MDD-Eval variants on the six dialogue evaluation benchmarks. Scores with p-values larger than 0.05 are underlined (indicating statistical insignificance). The best score for each benchmark is highlighted in bold. The ablation metrics include MDD-T, MDD-C and MDD-CM, which refer to the teacher model, the student model optimized with only $\mathcal{L}_{CE}$, and the student model optimized with both $\mathcal{L}_{MLM}$ and $\mathcal{L}_{CE}$ respectively. MDD-S is the full student model optimized with all three losses.}
\label{tab:correlation-main}
\end{table*}

\bigskip
\noindent\textbf{Main Correlation Results}
\text{ } Table~\ref{tab:correlation-main} presents the Spearman correlation scores of baseline evaluation metrics, the proposed MDD-Eval metric, and its ablation versions, across six dialogue evaluation benchmarks. For each MDD-Eval variant, we train the model five times with different random seeds and report the average results across the five runs. It can be observed that the full student model, MDD-S, performs generally well across all the evaluation benchmarks with an average Spearman correlation score of 0.476. MDD-S outperforms all the state-of-the-art model-based evaluation metrics. Remarkably, it outperforms the best baseline, DEB, by roughly 7\% in absolute terms. This confirms that MDD-Eval is a robust framework for the multi-domain dialogue evaluation task.

\bigskip
\noindent\textbf{Ablation Study}
\text{ } To better understand the influence of each component of MDD-S. The results w.r.t three ablation versions, MDD-T, MDD-C, and MDD-CM, are presented in Table~\ref{tab:correlation-main}. MDD-T is the teacher model, which is trained only on the single-domain human-annotated dataset, DailyDialog++. It performs well on the DailyDialog-Eval and Empathetic-Eval benchmarks,  which are close in domain w.r.t its training data source, compared to the baselines. This confirms our statement in Section~\ref{sec:intro} that learning from humans is an effective approach to equip ADMs with a rating skill to discriminate responses of varying quality.

MDD-C brings significant performance improvement over MDD-T (7.1\% Spearman correlation score). Note that MDD-C is learned with the vanilla self-training setup without consistency regularization and domain adaptation. The performance improvement showcases that the student model can generalize the rating skill of the teacher through the MDD-Data alone without any additional inductive bias.

MDD-CM brings a further improvement of 2.4\% Spearman correlation score. This confirms the usefulness of the self-supervised MLM objective in helping the student model to extract additional domain-specific knowledge. 

Finally, the full model MDD-S achieves the highest average Spearman correlation score of 0.476. This showcases the effectiveness of consistency regularization in our self-training setup.

\bigskip
\noindent\textbf{MDD-Eval vs uBERT-R} 
Unreferenced BERT-RUBER (uBERT-R) can be considered the fundamental representative of the recent family of ADMs based on self-supervised learning and pretrained language models. It can be observed that uBERT-R performs much worse than the MDD-Eval variants. There are two major reasons. First, uBERT-R acquires the rating skill to discriminate varying-quality responses in a self-supervised manner. The random sampling strategy adopted by uBERT-R is prone to introduction of false-negative and over-simplistic samples that can negatively impact the evaluation performance. The better performance of MDD-T than uBERT-R indicates that the human-designed sampling strategy is much more useful than the automatic random sampling scheme for equipping ADMs with the rating skill. Second, MDD-S generalizes the rating skill to multiple domains with a self-training framework in which additional inductive biases are incorporated, including mask language modeling and consistency regularization. The much better performance of MDD-S than uBERT-R showcases that semi-supervised learning is a promising option in improving dialogue evaluation performance compared to purely unsupervised learning.

\begin{figure*}[h!]
\centering
\includegraphics[width=0.70\textwidth]{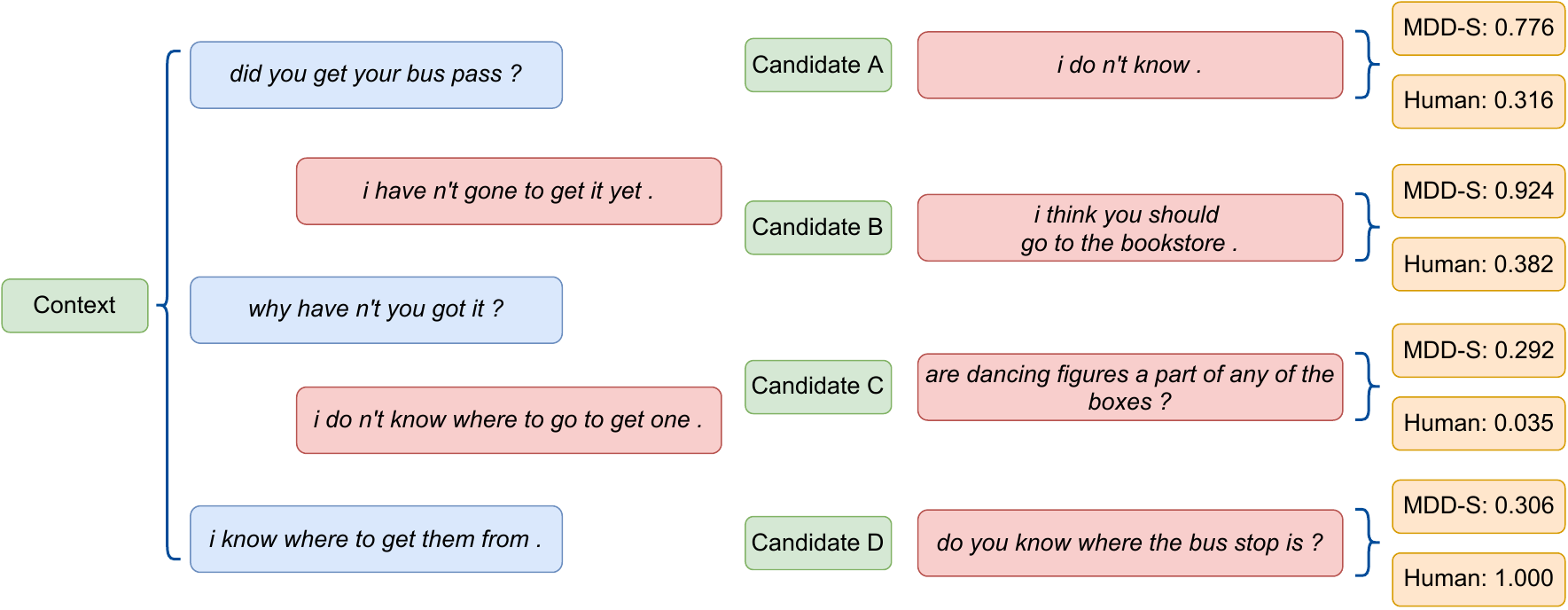} 
\caption{A case study to examine the limitations of MDD-Eval. The ordinal scores of both human and MDD-S are normalized to be within the [0, 1] range, and presented in the yellow box.}  
\label{error-example}
\end{figure*}

\bigskip
\noindent\textbf{MDD-Eval vs DEB} DEB and MDD-Eval variants are learned with the same classification task and their backbone model architectures are also similar. The only difference between MDD-T and DEB is that DEB is equipped with the general knowledge about dialogues across multiple domains through pretraining on 727M Reddit conversations with the MLM objective. As a result, DEB outperforms MDD-T by 3.5\% in terms of the average Spearman correlation score. This shows that pretraining on large-scale conversations is useful for the multi-domain dialogue evaluation task. However, DEB performs worse than MDD-S, which is trained only on 600K context-response pairs. The key difference between MDD-S and DEB is the generalization strategy. DEB adopts the pretrain-and-finetune paradigm whereas MDD-S adopts self-training. The more superior performance of MDD-S confirms that the self-training strategy is more effective and data-efficient.


\bigskip
\noindent\textbf{MDD-Eval vs Other Metrics } Even though GRADE, USL-H, USR, and D-score, have different training configurations, each of them have its unique strengths. Unlike uBERT-R, the four metrics have additional knowledge to generalize their evaluation skill. GRADE leverages Conceptnet Numberbatch~\citep{speer2017conceptnet}, which provides additional commonsense knowledge and topic information, to aid the self-supervised learning process. USR, USL-H and D-score consist of multiple model-based sub-metrics, and hence, they leverage more inductive biases for the task. It can be observed that MDD-S significantly outperforms the four state-of-the-art-metrics, confirming the effectiveness of the proposed self-training strategy for evaluation skill generalization. Since none of the current state-of-the-art metrics is explicitly designed to target the multi-domain dialogue evaluation problem, MDD-Eval helps bridge this gap.  

\bigskip
\noindent\textbf{Effects of Combining Data of Different Domains }There may be concern that some state-of-the-art metrics are trained on much less data or fewer dialogue domains compared to MDD-S. We presents the results w.r.t USL-H, USR, and GRADE in Table~\ref{tab:combined-correlation}. These three metrics are trained on a combined dataset, which contains the training data of all four dialogue corpora used to construct MDD-Data. We didn't include DEB (the best performing baseline) here, because DEB has already been pre-trained on large-scale Reddit conversations ($\sim$767M), and then finetuned on the high-quality DailyDialog++ dataset. We hypothesize that further finetuning DEB an mixed data will suffer from catastrophic forgetting. In addition, it can be observed that our MDD-C approach, which has similar model architecture and objective function as DEB, outperforms DEB on average, but performs worse compared to the final MDD-S metric. 

 It can be observed that simply combining dialogue data from different domains and training ADMs on the combined data in a self-supervised fashion don't bring robust performance for multi-domain dialogue evaluation. Hence, we need mechanism to filter undesirable data while keeping the ones useful to the evaluation task in order to construct a high-quality multi-domain dataset. MDD-Eval offers a simple, yet effective way to realize that.   

\begin{table}[!t]
\centering
\resizebox{0.7\linewidth}{!}{
\begin{tabular}{@{}lccccccc@{}}
\toprule
\multicolumn{2}{l}{Benchmarks}& \multicolumn{2}{c}{USR} & \multicolumn{2}{c}{USL-H} & \multicolumn{2}{c}{GRADE} \\ \midrule
\multicolumn{2}{l}{DailyDialog-Eval}& \multicolumn{2}{c}{0.491} & \multicolumn{2}{c}{0.358} & \multicolumn{2}{c}{0.485} \\
\multicolumn{2}{l}{Persona-Eval} & \multicolumn{2}{c}{0.174} &  \multicolumn{2}{c}{0.431} & \multicolumn{2}{c}{0.551} \\
\multicolumn{2}{l}{Topical-Eval} & \multicolumn{2}{c}{0.159} & \multicolumn{2}{c}{0.376} & \multicolumn{2}{c}{0.271} \\ 
\multicolumn{2}{l}{Empathetic-Eval} & \multicolumn{2}{c}{0.378} & \multicolumn{2}{c}{0.377} & \multicolumn{2}{c}{0.326} \\
\multicolumn{2}{l}{Movie-Eval}  & \multicolumn{2}{c}{0.505} & \multicolumn{2}{c}{0.515} & \multicolumn{2}{c}{0.559} \\
\multicolumn{2}{l}{Twitter-Eval} & \multicolumn{2}{c}{0.196} & \multicolumn{2}{c}{0.158} & \multicolumn{2}{c}{0.080} \\\midrule
\multicolumn{2}{l}{Average} & \multicolumn{2}{c}{0.317} & \multicolumn{2}{c}{0.369} & \multicolumn{2}{c}{0.379} \\
\bottomrule
\end{tabular}
}
\caption{Spearman correlation scores of USR, USL-H, and GRADE trained on the combined dataset.}
\label{tab:combined-correlation}
\end{table}

\bigskip
\noindent\textbf{Error Analysis } We can observe in Table~\ref{tab:correlation-main} that DEB outperforms MDD-S on Movie-Eval by a large margin. Similarly, D-score also outperforms MDD-S on Twitter-Eval by a large margin. We hypothesize this is because DEB has been pre-trained on 767M Reddit conversations (that could contain information about movies). In addition, we directly use a D-score checkpoint, which is trained on Twitter dialogues, for evaluating D-score's performance on Twitter-Eval. However, for MDD-S, the data distributions of Movie-Eval and Twitter-Eval are very different from its training datasets. This problem can be easily addressed by extend the MDD-data to both the movie, and the twitter domains.

To further analyze the limitations of MDD-Eval, we select a dialogue context and four candidate responses from the DailyDialog-Eval, and then, perform a case study on how MDD-S score the responses. The case study is presented in Figure~\ref{error-example}. Firstly, it can be observed that MDD-S provides a high score to a generic response, "I do n't know" while human annotators deem it inappropriate. Future work on MDD-Eval needs to consider modeling specificity. 

In addition, the metric focuses more on the neighbouring context, and struggles to capture key information in the longer context. For example, it doesn't recognize that the conversation is about "bus pass", which has little association with "book store" in candidate B. However, candidate B is somehow directive w.r.t its previous utterance (making a suggestion). Hence, MDD-S assigns a high score to candidate B. Future work may consider explicit modeling of speaker dependency, utterance dependency~\citep{zhang-etal-2021-dynaeval}, and the entity transition pattern within the dialogues.




\section{Conclusion}
\label{sec:conclusion}

We target the multi-domain dialogue evaluation problem and approach the problem with two research questions: (1) How can an ADM learn the rating skill to discriminate responses of varying quality? (2) How can the ADM acquire the general knowledge across different dialogue domains so as to generalize the evaluation skill? We propose MDD-Eval to address the two research questions. Specifically, a teacher evaluator is trained with human-annotated data to acquire the skill to distinguish good context-respons pairs from bad ones in a particular domain. Then, a new evaluator is trained with the teacher-annotated multi-domain data so as to generalizes the evaluation skill across multiple domains. Empirical results demonstrate that MDD-Eval is effective and robust for the multi-domain dialogue evaluation task.

\newpage
\section*{Acknowledgement}
We would like to thank all the reviewers for their constructive comments. This work is supported by Science and Engineering Research Council, Agency of Science, Technology and Research (A*STAR), Singapore, through the National Robotics Program under Human-Robot Interaction Phase 1 (Grant No. 192 25 00054);  Human Robot Collaborative AI under its AME Programmatic Funding Scheme (Project No. A18A2b0046); Robert Bosch (SEA) Pte Ltd under EDB’s Industrial Postgraduate Programme – II (EDB-IPP), project title: Applied Natural Language Processing; The work leading to these results is also part of the project GOMINOLA (PID2020-118112RB-C22) funded by MCIN/AEI/10.13039/501100011033 and project AMIC-PoC (PDC2021-120846-C42) funded by MCIN/AEI/10.13039/501100011033 and by “the European Union “NextGenerationEU/PRTR”.

\bibliography{main.bbl}

\newpage

\end{document}